\documentclass[11pt,a4paper]{article}
\usepackage[hyperref]{acl2020}
\usepackage{times}
\usepackage{latexsym}

\usepackage{graphicx}
\usepackage{multirow}
\usepackage{makecell}
\usepackage{subcaption}
\usepackage{tabularx}
\setcellgapes{5pt}

\usepackage{microtype}

\aclfinalcopy 

\title{Katecheo: A Portable and Modular System for Multi-Topic Question Answering}

\author{Shirish Hirekodi, Seban Sunny, \\
  \textbf{Leonard Topno, Alwin Daniel,} \\
  \textbf{Reuben Skewes, Stuart Cranney} \\
  CV Digital \\
  Sunshine Coast, Queensland, Australia \\
  {\tt ailab@cvglobal.co} \\\And
  Daniel Whitenack \\
  SIL International \\
  Dallas, TX, USA \\
  {\tt dan@datadan.io} \\}

\date{}

\begin{document}
\maketitle
\begin{abstract}
  We introduce a modular system that can be deployed on any Kubernetes cluster for question answering via REST API. This system, called Katecheo, includes three configurable modules that collectively enable identification of questions, classification of those questions into topics, document search, and reading comprehension. We demonstrate the system using publicly available knowledge base articles extracted from Stack Exchange sites. However, users can extend the system to any number of topics, or domains, without the need to modify any of the model serving code or train their own models. All components of the system are open source and available under a permissive Apache 2 License.
\end{abstract}

\section{Introduction}

When people interact with chatbots, smart speakers or digital assistants (e.g., Siri\footnote{\url{https://www.apple.com/siri/}}), one of their primary modes of interaction is information retrieval~\cite{Lovato:2015:STY:2771839.2771910}. Thus, those that build dialog systems often have to tackle the problem of question answering. 

Developers could support question answering using publicly available chatbot platforms, such as Watson Assistant\footnote{\url{https://www.ibm.com/cloud/watson-assistant/}} or DialogFlow\footnote{\url{https://dialogflow.com/}}. To do this, a user would need to program an intent for each anticipated question with various examples of the question and one or more curated responses. This approach has the advantage of generating high quality answers, but it is limited to those questions anticipated by developers. Moreover, the management burden of such a system might be prohibitive as the number of questions that need to be supported is likely to increase over time.

To overcome the burden of programming intents, developers might look towards more advanced question answering systems that are built using open domain question and answer data (e.g., from Stack Exchange or Wikipedia), reading comprehension models, and document search techniques. In particular, \citeauthor{chen-etal-2017-reading}~previously demonstrated a two step system, called DrQA, that matches an input question to a relevant article from a knowledge base and then uses a recurrent neural network (RNN) based comprehension model to detect an answer within the matched article. This more flexible method was shown to produce promising results for questions related to Wikipedia articles and it performed competitively on the SQuAD benchmark~\cite{rajpurkar-etal-2016-squad}. 

However, if developers want to integrate this sort of reading comprehension based methodology into their applications, how would they currently go about this? They would need to wrap pre-trained models in their own custom code and compile similar knowledge base articles at the very least. At the most, they may need to re-train reading comprehension models on open domain question and answer data (e.g., SQuAD) and/or implement their own knowledge base search algorithms. 

In this paper we present \emph{Katecheo}, a portable and modular system for reading comprehension based question answering that attempts to ease this development burden. The system provides a quickly deployable and easily extendable way for developers to integrate question answering functionality into their applications. Katecheo includes three configurable modules that collectively enable identification of questions, classification of those questions into topics, document search, and reading comprehension. The modules are tied together in a single inference graph that can be invoked via a REST API call. We demonstrate the system using publicly available knowledge base articles extracted from Stack Exchange sites\footnote{\url{https://stackexchange.com/sites\#}}. However, users can extend the system to any number of topics, or domains, without the need to modify the model serving code. All components of the system are open source and publicly available under a permissive Apache 2 License\footnote{\url{https://github.com/cvdigitalai/katecheo}}.

The rest of the paper is organized as follows. In the next section, we provide an overview of the system logic and its modules. In Section 3, we outline the architecture and configuration of Katecheo, including extending the system to an arbitrary number of topics. In Section 4, we report some results using public knowledge base articles. Then in conclusion, we summarize the system, its applicability, and future development work.
  
\section{System Overview}

\begin{figure}
  \makebox[\columnwidth]{\includegraphics[width=\columnwidth]{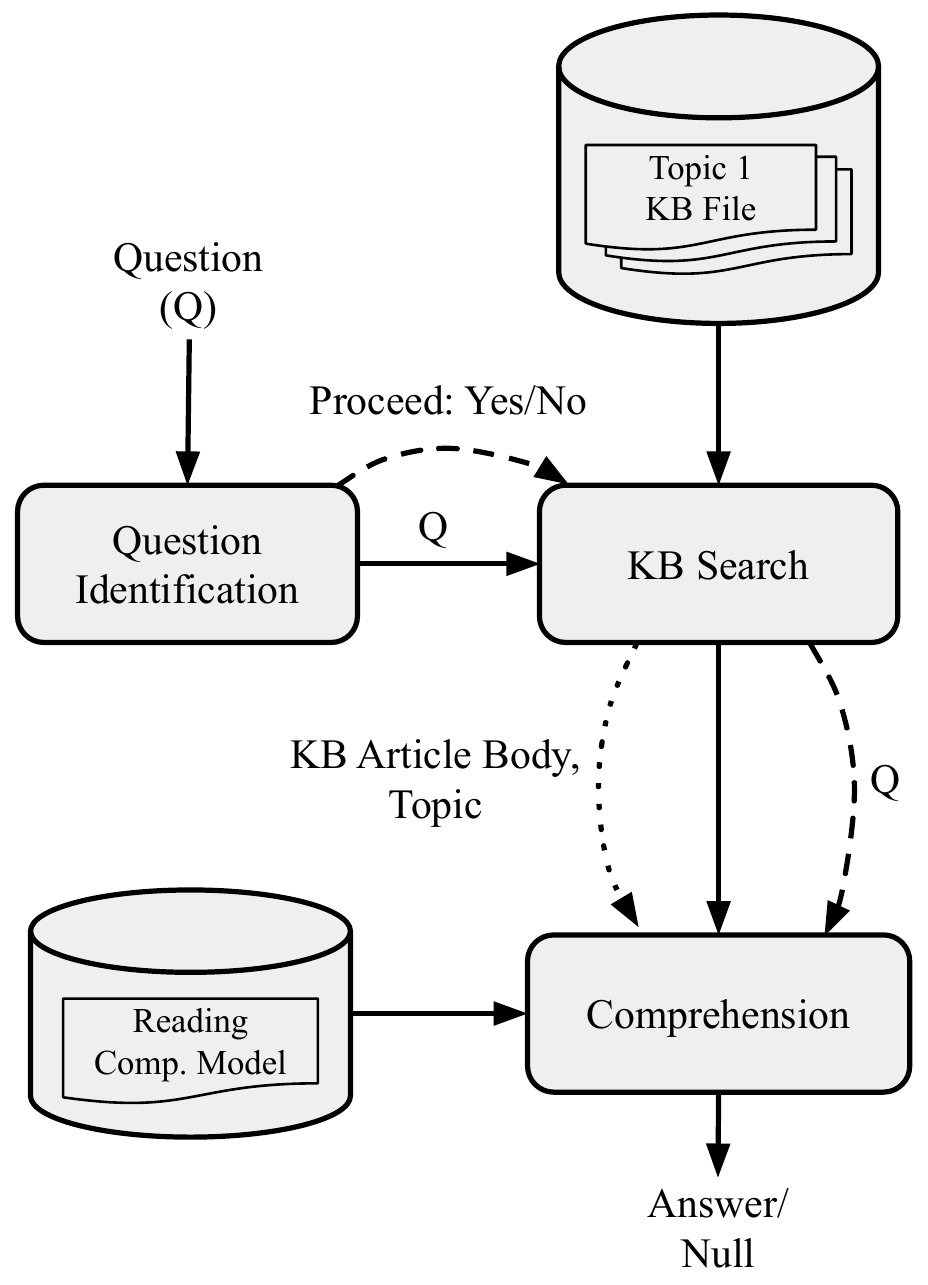}}
  \caption{The overall processing flow in Katecheo. \emph{Q} represents the input question text, the dashed lines represent metadata passed between modules indicating whether the next module should proceed with processing, and the cylinders represent various data inputs to the modules.}
  \label{flow}
\end{figure}

Katecheo is partially inspired by the work of~\citeauthor{chen-etal-2017-reading} on DrQA. That previously developed method has two primary phases of question answering: document retrieval and reading comprehension. Together these functionalities enable open domain question answering. However, many dialog systems are not completely open domain. For example, developers might want to create a chatbot that has targeted conversations about restaurant reservations and movie times. It would be advantageous for such a chatbot to answer questions about food and entertainment, but the developers might not want to allow the conversation to stray into other topics.

One of the goals of Katecheo was to create a question answering system that is more flexible than those relying on curated responses while remaining more targeted than a completely open domain question answering system. The system includes document retrieval (or what we refer to as ``knowledge base search'') and reading comprehension, but only within sets of curated knowledge base articles each corresponding to a particular topic (e.g., food and/or entertainment). 

When a question text is input into the Katecheo system, it is processed through three modules: (1) question identification, (2) knowledge base search, and (4) reading comprehension. This overall logic is depicted in Figure~\ref{flow}.

\subsection{Question Identification}

The first module in Katecheo, question identification, determines if the input text (labeled \emph{Q} in Figure~\ref{flow}) is actually a question. In our experience, users of dialog systems provide a huge number of unexpected inputs. Some of these unexpected inputs are questions and some are just statements. Before going to the trouble of matching a knowledge base article and generating an answer, Katecheo completes this initial step to ensure that the input is a question. If the input is a question, the question identification module (henceforth the ``question identifier") passes a positive indication/flag to the next module indicating that it should continue processing the positively identified question. Otherwise, it passes an error flag that ends the processing.

The question identifier uses a rule-based approach to question identification. As suggested in~\citeauthor{Li:2011:QIT:2063576.2063996}, we utilize the presence of question marks and 5W1H words to determine if the input is a question. To test this question identification technique, we extracted 3000 questions from the SQuAD dev data set~\cite{rajpurkar-etal-2018-know}. We also extracted an equal number of statements from the contexts in the SQuAD dev data set using sentence segmentation. The overall accuracy of our rule-based approach to question identification on this test data is 95.1\%, and the question identifier tends to predict very few false negatives as compared to false positives (as shown in Table~\ref{confusion}).

\begin{table}
\begin{center}
{    
\makegapedcells
\begin{tabular}{cc|cc}
\multicolumn{2}{c}{}
            &   \multicolumn{2}{c}{Predicted} \\
    &       &   Question &   Statement              \\ 
    \cline{2-4}
\multirow{2}{*}{\rotatebox[origin=c]{90}{Actual}}
    & Question   & 5975   & 25                 \\
    & Statement    & 530    & 5470                \\ 
    \cline{2-4}
\end{tabular}
}
\end{center}
\caption{\label{confusion} Confusion matrix detailing the performance of the rule-based question identification technique utilized in the question identifier module of Katecheo. }
\end{table}

\subsection{Knowledge Base Search}

Assuming the input text is identified as a question, a search is made to match the question with an appropriate knowledge base article from a set of user supplied knowledge base articles, which set corresponds to a user supplied topic. In this way, the system also assigns the question to a particular topic. The matched article will be utilized in the next stage of processing to generate an answer.

The user supplies one or more sets of knowledge base articles, where each set is tagged with a topic name. These sets of articles are formatted into JSON format with an array containing the knowledge base articles. Each article in the array has a title field, a body field, and an article ID field. 

In the knowledge base search module of Katecheo (henceforth the ``KB Search" module), articles and questions are compared using TF-IDF vectors and cosine similarity. However, given the potential for multiple sets of knowledge base articles (corresponding to multiple topics), these articles can be searched in (1) a segmented manner (where each set of knowledge base articles is searched in isolation), or (2) a combined manner (where the sets of knowledge base articles are concatenated and one search is performed on the concatenated sets). 

To investigate the implications of using segmented or combined searching, we compiled a set of 6,110 Medical Sciences related articles from the WebMD site\footnote{\url{https://www.webmd.com/}} and a set of 6,118 Christianity-related articles from the GotQuestions website\footnote{\url{https://www.gotquestions.org/}}. We then annotated 50 Christianity related questions and 50 Medical Sciences related questions, where these 100 questions could be answered using a tagged one of the compiled articles.  Finally, we performed segmented and combined searching for each of the 100 questions using the compiled articles and a smaller subset of the compiled articles. The results are shown in Table~\ref{search}.

\begin{table}
\begin{center}
{    
\makegapedcells
\begin{tabular}{cc|cc}
\multicolumn{2}{c}{}
            &   \multicolumn{2}{c}{3,000 articles from each set} \\
    &       &   Christianity &   Med. Sciences              \\ 
    \cline{2-4}
\multirow{2}{*}{\rotatebox[origin=c]{90}{Method}}
    & Segmented    & 36\%   & 60\%                 \\
    & Combined     & \textbf{42\%}    & \textbf{64\%}                \\ 
    \cline{2-4}
\end{tabular}
}
{    
\makegapedcells
\begin{tabular}{cc|cc}
\multicolumn{2}{c}{}
            &   \multicolumn{2}{c}{} \\
\multicolumn{2}{c}{}
            &   \multicolumn{2}{c}{All 12,228 articles} \\
    &       &   Christianity &   Med. Sciences              \\ 
    \cline{2-4}
\multirow{2}{*}{\rotatebox[origin=c]{90}{Method}}
    & Segmented    & 36\%   & 60\%                 \\
    & Combined     & 30\%    & 58\%                \\ 
    \cline{2-4}
\end{tabular}
}
\end{center}
\caption{\label{search} Accuracy values for matching a question with the most relevant knowledge base article using either a combined or segmented search strategy. }
\end{table}

The most accurate method of searching through articles for smaller numbers of articles (e.g., 6000 or less in total) is the combined search method, while larger sets of articles seem to benefit from a segmented search strategy. Given that Katecheo users are expected to upload their own knowledge base articles for particular topics, we decided to employ the combined search strategy and assume that users will be able to curate their knowledge base articles into appropriately sized sets. In the future we anticipate extending Katecheo to optionally support both segmented and combined search depending on the size of input knowledge bases. 

In addition to matching a question to an article and topic, the KB search module of Katecheo also uses a cosine similarity threshold value to filter out completely off topic questions (i.e., those that are dissimilar to one of the user supplied topics). This threshold helps Katecheo achieve the goal of being flexible while not completely open domain, which is often a goal of chatbot and assistant applications. The threshold value is optionally configurable by the user and can range from 0 to 1. Higher threshold values of cosine similarity will result is a very strict enforcement of questions aligning with the supplied topics, and lower threshold values result in more tolerance.

\begin{figure}
  \makebox[\columnwidth]{\includegraphics[width=\columnwidth]{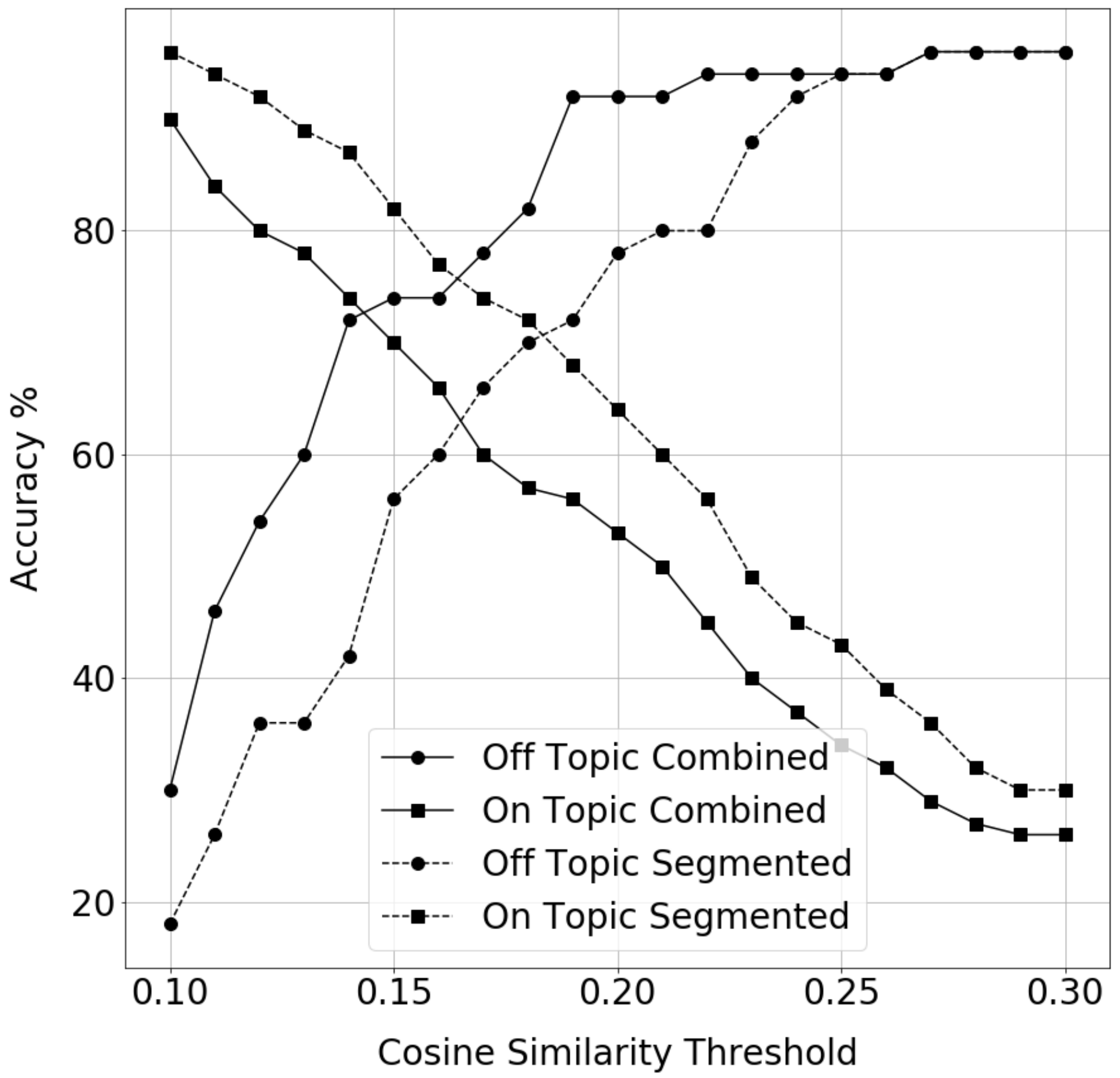}}
  \caption{Accuracy of identifies on-topic and off-topic questions as the cosine similarity threshold is varied between 0.10 and 0.30. The solid lines correspond to a combined search strategy and the dashed lines correspond to a segmented search strategy.}
  \label{threshold}
\end{figure}

To choose an appropriate default threshold value for cosine similarity in the TF-IDF based search, we performed another experiment using the WebMD and GotQuestions data. In this case, we added an additional 50 off topic questions that were dissimilar to the Christianity and Medical Science topics, such as questions about sports or movies. We then varied the cosine similarity threshold while observing the number of correctly identified on-topic questions and off-topic questions. The results of this analysis are shown in Figure~\ref{threshold}. Because we will be using a combined approach to search and wanted to filter out the majority of off topic questions, we use 0.15 as our default cosine similarity threshold.

\subsection{Reading Comprehension}

The final module of the Katecheo system is the reading comprehension (or just ``comprehension") module. This module takes as input the original input question plus the matched knowledge base article body text and uses a reading comprehension model to select an appropriate answer from within the article.

Users of Katecheo can configure the system to utilize one of a two different pre-trained, reading comprehension models. In the current release, users can choose between: (1) a Bi-Directional Attention Flow, or BiDAF, model~\cite{Seo2017BidirectionalAF}; and (2) a large BERT~\cite{devlin2018bert} whole-word model fine-tuned on SQuAD 1.0~\cite{rajpurkar-etal-2018-know}. We are using a pre-trained version of BiDAF available in the AllenNLP~\cite{Gardner2017AllenNLP} library and the pre-trained version of BERT available in the transformers library~\cite{Wolf2019HuggingFacesTS}. By default Katecheo will use the BERT model.

\section{Architecture and Configuration}

\begin{figure}
  \makebox[\columnwidth]{\includegraphics[width=\columnwidth]{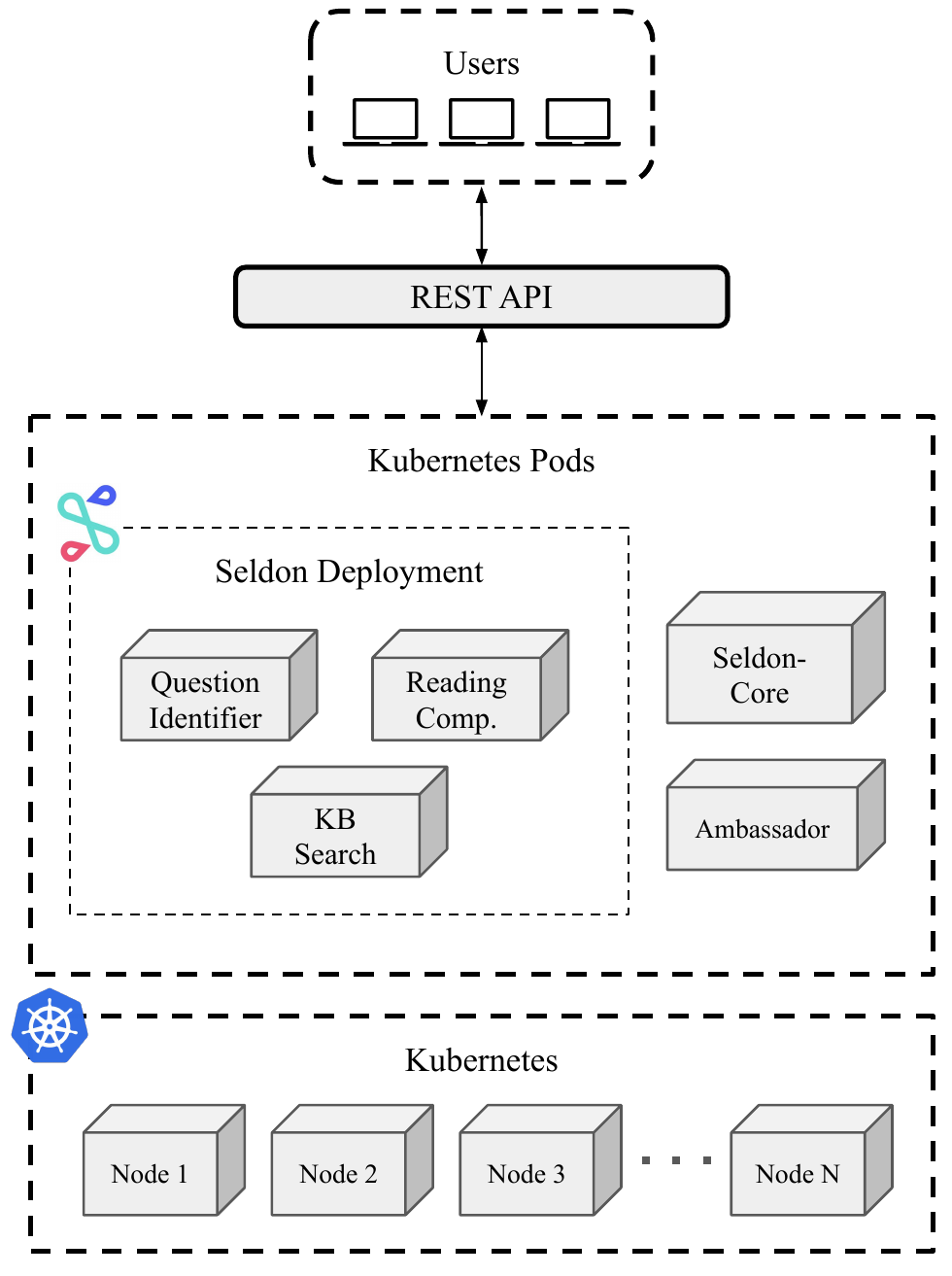}}
  \caption{The overall Katecheo architecture. Each \emph{node} in Kubernetes may be a cloud instance or on-premise machine.}
  \label{kube}
\end{figure}

All three of the Katecheo modules are containerized with Docker~\cite{Merkel:2014:DLL:2600239.2600241} and are deployed as pods on top of Kubernetes~\cite{Hightower:2017:KUR:3175917} (see Figure~\ref{kube}). In this way, Katecheo is completely portable to any standard Kubernetes cluster including hosted versions in AWS, GCP, Digital Ocean, Azure, etc. and on-premises version that use vanilla Kubernetes, OpenShift, CaaS, etc.

To provide developers with a familiar interface to the question answering system, we provide a REST API interface. Developers can call Katecheo via a single endpoint with ingress to the system provided by Ambassador\footnote{\url{https://github.com/datawire/ambassador}}, a Kubernetes-native API Gateway. 

Seldon-core\footnote{\url{https://github.com/SeldonIO/seldon-core}} is used to simplify the routing between the three modules, create the REST API, and manage deployments. To create the Seldon deployment of the three modules, as depicted in Figure~\ref{kube}, we: (1) create a Python class for each module that contains standardized Seldon-specified methods and that loads the various models for making predictions; (2) wrap that Python class in a standard, containerized Seldon model server using a public Seldon Docker image and s2i~\footnote{\url{https://github.com/openshift/source-to-image}}; (3) push the wrapped Python code to DockerHub~\footnote{\url{https://hub.docker.com/}}; (4) create a Seldon inference graph that links the modules in a Directed Acyclic Graph (DAG); and (5) deploy the inference graph to Kubernetes. After all of these steps are complete, a single REST API endpoint is exposed. When a user calls this single API endpoint the Seldon inference graph is invoked and the modules are executed using the specified routing logic.

To specify the topic names and topic knowledge base JSON files (as mentioned in reference to Figure~\ref{flow}), the user need only fill out a JSON configuration file template listing the topic name and URL link for each knowledge base file. These could be static files or served via a separate API as appropriate. Once the configuration file is created, a Bash deploy script can be executed to automatically deploy all of the Katecheo modules to a Seldon-enabled Kubernetes cluster.

\section{Example Usage}

\begin{table*}[t!]
\centering
\small
\makegapedcells
\caption{\label{results} Example inputs, outputs, matched topics and matched articles from a Katecheo system deployed to provide question answering on two topics, Medical Sciences (Panel A) and Christianity (Panel B). Katecheo was configured in this case with knowledge base articles from stack exchange sites and a cosine similarity threshold value of 0.15.}
\subcaption*{Panel A: On-topic, Medical questions with relevant articles in the knowledge base.}
\begin{tabular}{|p{2.5cm}|p{2cm}|p{3cm}|p{3cm}|p{3cm}|}
\hline \bf Question & \bf Matched Topic & \bf Matched Article Title & \bf Answer (BERT) & \bf Answer (BiDAF) \\ \hline
Why do we get cold sores? & Med. Sciences & Cold sores: why do we get them on the lips? & In times of stress, fever, illness or even over exposure to sunlight, & an infection with the type 1 or Type 2 herpes simplex virus \\
\hline
How should you treat people with high risk factors for coronary heart disease? & Med. Sciences & Can Prediabetes cause coronary heart disease? & aspirin and/or statins & aspirin and/or statins \\
\hline
What is the best way to reduce pain and swelling in a knee joint? & Med. Sciences & Pain in knee joint & Applying cold compresses & Applying cold \\
\hline
Which would kill you first, hypothermia or frost bite? & None & None & None & None \\
\hline
\end{tabular}
\bigskip
\subcaption*{Panel B: On-topic, Christianity questions with relevant articles in the knowledge base.}
\begin{tabular}{|p{2.5cm}|p{2cm}|p{3cm}|p{3cm}|p{3cm}|}
\hline \bf Question & \bf Matched Topic & \bf Matched Article Title & \bf Answer (BERT) & \bf Answer (BiDAF) \\ \hline
What does LDS theology and official teaching say about who goes to Hell? & Christianity & Who goes to hell in LDS theology? & all who have died without the knowledge of truth & rejected the truth will \\
\hline
What is the Messianic Secret? & Christianity & Jesus concealing his identity & a prohibition to make known the messianic character of Jesus. & a prohibition to make known the messianic character of Jesus \\
\hline
What did Bart Ehrman say about Church scribes and the Bible? & Christianity & How do apologists defend against Bart Ehrman's arguments that Church scribes corrected and changed the Bible to fit their theology? & corrected and changed the Bible to fit their theology? & to fit their theology \\
\hline
What is the biblical basis for God being omnipresent? & None & None & None & None \\
\hline
\end{tabular}
\end{table*}

We demonstrated the utility of Katecheo by deploying the system for question answering in two topics, \emph{Medical Sciences} and \emph{Christianity}. These topics are diverse enough that they would warrant different curated sets of knowledge base articles, and we can easily retrieve knowledge base articles for each of these subjects from the Medical Sciences\footnote{\url{https://medicalsciences.stackexchange.com/}} and Christianity\footnote{\url{https://christianity.stackexchange.com/}} Stack Exchange sites, respectively. Our extracted and formatted are publicly available and linked in the Katecheo GitHub repository.

Example inputs and outputs of the deployed system are included in Table~\ref{results}. As can be seen, the system is able to match many questions with an appropriate topic and article and subsequently generate an answer using the implemented comprehension models. Not all of the answers would fit into conversational question answering in terms of naturalness, but others show promise. 

In our experience, the number and curation of the knowledge base articles directly influences the performance of the system more than any other single variable. The KB search module in our example deployment consistently matched more \emph{Medical Sciences} questions to relevant articles than it did \emph{Christianity} questions. The \emph{Christianity} articles seem to be much more similar to one another in terms of vocabulary than the \emph{Medical Sciences} articles. This vocabulary overlap creates problems for the KB Search module and may reduce the cosine similarity scores for any \emph{Christianity} related question. In turn, reduced similarity scores cause many relevant questions to be classified as off-topic. 

We recommend careful curation of knowledge base articles for input to Katecheo. Smaller topical knowledge bases with high quality, diverse articles will result in better question-to-article matching than large quantities of articles acquired indiscriminately. 

\section{Conclusions}

In conclusion, Katecheo is a portable and modular system for reading comprehension based question answering. It is portable because it is built on cloud native technologies (i.e., Docker and Kubernetes) and can be deployed to any cloud or on-premise environment. It is modular because it is composed of three configurable modules that collectively enable identification of questions, classification of those questions into topics, a search of knowledge base articles, and reading comprehension.

Initial usage of the system indicates that it provides a flexible and developer friendly way to enable question answering functionality for multiple topics or domains via REST API. That being said, the current configurations of Katecheo performs best with smaller sets of knowledge base articles (6000 or less). We plan to overcome this limitation by upgrading the TF-IDF based document search implementation using, e.g., n-grams and/or more sophisticated language models. In addition, future development of Katecheo will include features that allow users to (i) dynamically adjust the cosine similarity threshold and reading comprehension model, (ii) utilize other reading comprehension models or even custom reading comprehension models, and (iii) return multiple answers or an ensembled answer generated from multiple comprehension models.

The complete source code, configuration information, deployment scripts, and examples for Katecheo are available at \url{https://github.com/cvdigitalai/katecheo}. A screencast demonstration of Katecheo is available at \url{here}. A example Streamlit\footnote{\url{https://www.streamlit.io/}} app allowing user-friendly testing of a Katecheo deployment is available at \url{here}. 


\bibliography{anthology,acl2020}
\bibliographystyle{acl_natbib}

\end{document}